\documentclass[11pt,a4paper]{article}

\pdfoutput=1

\usepackage[hyperref]{emnlp2018}
\usepackage[utf8]{inputenc}
\usepackage{times}
\usepackage{latexsym}
\usepackage{amsmath,amssymb,mathtools}
\usepackage{linguex}
\usepackage{todonotes}
\usepackage{enumitem}
\usepackage{booktabs}

\usepackage{url}

\usepackage{aswhitedefs}

\aclfinalcopy 
\def\aclpaperid{1369} 

\newcommand{\figref}[1]{Figure \ref{#1}}
\newcommand{\tabref}[1]{Table \ref{#1}}

\title{Lexicosyntactic Inference in Neural Models}

\author{Aaron Steven White\\University of Rochester \And 
        Rachel Rudinger\\Johns Hopkins University \AND
        Kyle Rawlins\\Johns Hopkins University \And
        Benjamin Van Durme\\Johns Hopkins University
        }

\date{}

\begin{document}
\maketitle
\begin{abstract}
We investigate neural models' ability to capture \textit{lexicosyntactic inferences}: inferences triggered by the interaction of lexical and syntactic information. We take the task of event factuality prediction as a case study and build a factuality judgment dataset for all English clause-embedding verbs in various syntactic contexts. We use this dataset, which we make publicly available, to probe the behavior of current state-of-the-art neural systems, showing that these systems make certain systematic errors that are clearly visible through the lens of factuality prediction.
\end{abstract}

\renewcommand{\firstrefdash}{}

\setlength{\Exlabelsep}{.2em}
\setlength{\Extopsep}{.2\baselineskip}

\setlength{\SubExleftmargin}{1em}

\section{Introduction}
\label{sec:introduction}

\vspace{-1mm}

The formal semantics literature has long been concerned with the complex array of inferences that different open class lexical items trigger \citep{kiparsky_fact_1970,karttunen_implicative_1971,karttunen_observations_1971,horn_semantic_1972,karttunen_conventional_1979,heim_presupposition_1992,simons_conversational_2001,simons_observations_2007,simons_what_2010,abusch_lexical_2002,abusch_presupposition_2010,gajewski_neg-raising_2007,anand_epistemics_2013,anand_factivity_2014}. For example, why does \ref{ex:hypothesisbelieve} give rise to the inference \ref{ex:inferencebelieve}, while the structurally identical \ref{ex:hypothesisknow} triggers the inference \ref{ex:inferenceknow}?   


\ex. 
\a. Jo doesn't believe that Bo left. \label{ex:hypothesisbelieve}
\b. Jo doesn't know that Bo left. \label{ex:hypothesisknow}

\ex. 
\a. Jo believes that Bo didn't leave. \label{ex:inferencebelieve}
\b. Bo left. \label{ex:inferenceknow}
\c. Bo didn't leave. \label{ex:inferencerememberto}

A major finding of this literature is that lexically triggered inferences are conditioned by surprising aspects of the syntactic context that a word occurs in. For example, while \ref{ex:rememberthat}, \ref{ex:notrememberthat}, and \ref{ex:rememberto} trigger the inference \ref{ex:inferenceknow}, \ref{ex:notrememberto} triggers the inference \ref{ex:inferencerememberto}. 

\ex. \label{ex:rememberfinite}
\a. Jo remembered that Bo left. \label{ex:rememberthat}
\b. Jo didn't remember that Bo left. \label{ex:notrememberthat}

\ex. \label{ex:remembercontrol}
\a. Bo remembered to leave. \label{ex:rememberto}
\b. Bo didn't remember to leave. \label{ex:notrememberto}



\noindent Accurately capturing such interactions -- e.g. between clause-embedding verbs, negation, and embedded clause type -- is important for any system that aims to do general natural language inference (\citealt{maccartney_phrase-based_2008} \textit{et seq}; cf. \citealt{dagan_pascal_2006}) or event extraction (see \citealt{grishman_message_1996} \textit{et seq}), and it seems unlikely to be a trivial phenomenon to capture, given the complexity and variability of the inferences involved \citep[see, e.g.,][on implicatives]{karttunen_simple_2012,karttunen_you_2013,karttunen_chameleon-like_2014,van_leusen_accommodation_2012,white_factive-implicatives_2014,baglini_implications_2016,nadathur_causal_2016}.

In this paper, we investigate how well current state-of-the-art neural systems for a subtask of general event extraction -- event factuality prediction \citep[EFP;][]{nairn_computing_2006,sauri_factbank:_2009,sauri_are_2012,de_marneffe_did_2012,lee_event_2015,stanovsky_integrating_2017, rudinger_neural_2018} -- capture inferential interactions between lexical items and syntactic context -- \textit{lexicosyntactic inferences} -- when trained on current event factuality datasets. Probing these particular systems is useful for understanding neural systems' behavior more generally because (i) the best performing neural models for EFP \citep{rudinger_neural_2018} are simple instances of common baseline models; and (ii) the task itself is relatively constrained.  

To do this, we substantially extend the MegaVeridicality1 dataset \citep{white_role_2018} to cover all English clause-embedding verbs in a variety of the syntactic contexts covered by recent psycholinguistic work \citep{white_computational_2016}, and we use the resulting dataset -- MegaVeridicality2 -- to probe these models' behavior.  We focus on clause-embedding verbs because they show effectively every possible patterning of lexicosyntactic inference \citep{karttunen_simple_2012}. 

We discuss three findings: (i) Tree biLSTMs (T-biLSTMs) are better able to correctly predict lexicosyntactic inferences than linear-chain biLSTMs (L-biLSTMs); (ii) L-biLSTMs and T-biLSTMs capture different lexicosyntactic inferences, and thus ensembling their predictions can reliably improve performance; and (iii) even when ensembled, these models show systematic errors -- e.g. performing well when the polarity of the matrix clause matches the polarity of the true inference, but poorly when these polarities mismatch. 




We furthermore release MegaVeridicality2 at \href{http://megaattitude.io}{MegaAttitude.io} as a benchmark for probing the ability of neural systems -- whether for factuality prediction or for general natural language inference -- to capture lexicosyntactic inference.


\vspace{-1mm}

\section{Data collection}
\label{sec:datacollection}

\vspace{-1mm}

We substantially extend the MegaVeridicality1 dataset \citep{white_role_2018}, which contains factuality judgments for all English clause-embedding verbs that take tensed subordinate clauses. In \citeauthor{white_role_2018}'s annotation protocol, all verbs that are grammatical with such subordinate clauses -- based on the MegaAttitude dataset \citep{white_computational_2016} -- are slotted into contexts either like \ref{ex:finiteintrans} or \ref{ex:finitetrans}, depending on whether they take a direct object or not.

\ex. 
\a. Someone \{knew, didn't know\} that a particular thing happened. \label{ex:finiteintrans}
\b. Someone \{was, wasn't\} told that a particular thing happened. \label{ex:finitetrans}

For each sentence generated in this way, 10 different annotators are asked to answer the question \textit{did that thing happen?}: \textit{yes}, \textit{maybe or maybe not}, \textit{no}.

There are two important aspects of these contexts to note. First, all lexical items besides the embedding verbs are semantically bleached to ensure that the measured lexicosyntactic inferences are only due to interactions between the embedding predicate -- e.g. \textit{know} or \textit{tell} -- and the syntactic context. Second, the matrix polarity -- i.e. the presence or absence of \textit{not} as a direct dependent of the embedding verb -- is manipulated to create two sentences for each verb-context pair.

Our extension, MegaVeridicality2, includes judgments for a variety of infinitival subordinate clause types, exemplified in \ref{ex:infinitival}.\footnote{We also explicitly manipulate two aspects of the subordinate clause in our extension of the MegaVeridicality dataset: (i) how NP embedded subjects are introduced; and (ii) whether the embedded clause contains an eventive predicate (\textit{do}, \textit{happen}) or a stative predicate (\textit{have}). See Appendix \ref{sec:datacollectionsupplement} for details on the reasoning behind these manipulations.} We investigate infinitival clauses because they can give rise to different lexicosyntactic inferences than finite subordinate clauses -- e.g. compare \ref{ex:rememberfinite} and \ref{ex:remembercontrol}.

\ex. \label{ex:infinitival}
\a. Someone \{needed, didn't need\} for a particular thing to happen. \label{ex:forto} 
\b. Someone  \{wanted, didn't want\} a particular person to do, have a particular thing.\label{ex:aci_eventive}
\b. Someone  \{wanted, didn't want\} a particular person to have a particular thing.\label{ex:aci_stative}
\b. A particular person \{was, wasn't\} overjoyed to do a particular thing. \label{ex:infinitivalpassive_eventive}
\b. A particular person \{was, wasn't\} overjoyed to have a particular thing. \label{ex:infinitivalpassive_stative}
\b. A particular person \{managed, didn't manage\} to do a particular thing. \label{ex:control_eventive}
\b. A particular person \{managed, didn't manage\} to have a particular thing. \label{ex:control_stative}

For each sentence, we also collect judgments from 10 different annotators, using the same question as \citeauthor{white_role_2018} for context \ref{ex:forto} and modified questions for contexts \ref{ex:aci_eventive}-\ref{ex:control_stative}: \textit{did that person do that thing?} for \ref{ex:aci_eventive}, \ref{ex:infinitivalpassive_eventive}, and \ref{ex:control_eventive}; and \textit{did that person have that thing?} for for \ref{ex:aci_stative}, \ref{ex:infinitivalpassive_stative}, and \ref{ex:control_stative}. \tabref{tab:verbcount} shows the number of verb types for each syntactic context. With the polarity manipulation, this yields a total of 3,938 sentences.

To build a factuality prediction test set from these sentences, we combine MegaVeridicality1 with our dataset and replace each instance of \textit{a particular person} or \textit{a particular thing} with \textit{someone} or \textit{something} (respectively). Then, following \citeauthor{white_role_2018}, we normalize the 10 responses for each sentence to a single real value using an ordinal mixed model-based procedure.  We refer to the resulting dataset as MegaVeridicality2.

\begin{table}
\scriptsize
\centering
\begin{tabular}{lrrl}
\toprule
\textbf{Syntactic context} &  \textbf{\# verbs} & \textbf{\# sents} & \textbf{Ex.} \\
\midrule
NP \_ed that S         &       375 & 750 & \ref{ex:finiteintrans}\\
NP was \_ed that S     &       169 & 338 & \ref{ex:finitetrans} \\
\midrule
NP \_ed for NP to VP   &       184 & 368 & \ref{ex:forto} \\
NP \_ed NP to VP[+ev]  &       197 & 394 & \ref{ex:aci_eventive}  \\
NP \_ed NP to VP[-ev]  &       128 & 256 & \ref{ex:aci_stative} \\
NP was \_ed to VP[+ev] &       278 & 556 & \ref{ex:infinitivalpassive_eventive}  \\
NP was \_ed to VP[-ev] &       256 & 512 & \ref{ex:infinitivalpassive_stative} \\
NP \_ed to VP[+ev]     &       217 & 434 & \ref{ex:control_eventive} \\
NP \_ed to VP[-ev]     &       165 & 330 & \ref{ex:control_stative} \\
\midrule
Total                  &       1,969 & 3,938 & \\
\bottomrule
\end{tabular}
\vspace{-1mm}
\caption{\small Contexts and number of verbs for which annotations were collected: S = \textit{something happened}, NP = \textit{someone}, VP = \textit{happen}, VP[+ev] = \textit{do something}, VP[-ev] = \textit{have something}. First two rows: MegaVeridicality1. All rows: MegaVeridicality2. The number of sentences is always twice the number of verbs, since matrix polarity is manipulated.}
\label{tab:verbcount}
\vspace{-5mm}
\end{table}

\section{Model and evaluation}
\label{sec:model}

\vspace{-1mm}

We use MegaVeridicality2 to evaluate the performance of three state-of-the-art neural models of event factuality \citep{rudinger_neural_2018}: a linear-chain biLSTM (L-biLSTM), a dependency tree biLSTM (T-biLSTM), and a hybrid biLSTM (H-biLSTM) that ensembles the two. To predict the factuality of the event referred to by a particular predicate, these models pass the output state of the biLSTM at that predicate  through a two-layer regression. In the case of the H-biLSTM, the output state of both the L- and T-biLSTMs are simply concatenated and passed through the regression.\footnote{See Appendix \ref{sec:modelsupplement} for further details.}

Following the multi-task training regime described by \citet{rudinger_neural_2018}, we train these models on four standard factuality datasets -- FactBank \citep{sauri_factbank:_2009,sauri_are_2012}, UW \citep{lee_event_2015}, MEANTIME \citep{minard_meantime_2016}, and UDS \citep{white_universal_2016,rudinger_neural_2018} -- with tied biLSTM weights but regression parameters specific to each dataset. We then use these trained models to predict the factuality of the embedded predicate in our dataset.

To understand how much of these models' performance on our dataset is really due to a correct computation of lexicosyntactic inferences, we also generate predictions for the sentences in our dataset with the embedding verbs UNKed. In this case, the model can rely only on the syntactic context surrounding the predicate to make its inferences. We refer to the models with lexical information as the LEX models and the ones without lexical information as the UNK models.

Each model produces four predictions, corresponding to the four different datasets it was trained on. We consider three different ways of ensembling these predictions using a cross-validated ridge regression: (i) ensembling the four predictions for each specific model (LEX or UNK); (ii) ensembling the predictions for the LEX version of a particular model with the UNK version of that same model (LEX+UNK); and (iii) ensembling the predictions across all models (LEX, UNK, or LEX+UNK). Each ensemble is evaluated in a 10-fold/10-fold nested cross-validation \citep[see][]{cawley_over-fitting_2010}.  In each iteration of the outer cross-validation, a 10\% test set is split off, and a 10-fold cross-validation to tune the regularization is conducted on the remaining 90\%.

\vspace{-1mm}

\section{Results}
\label{sec:results}

\vspace{-1mm}

\begin{figure}[t]
\centering
\includegraphics[scale=.5]{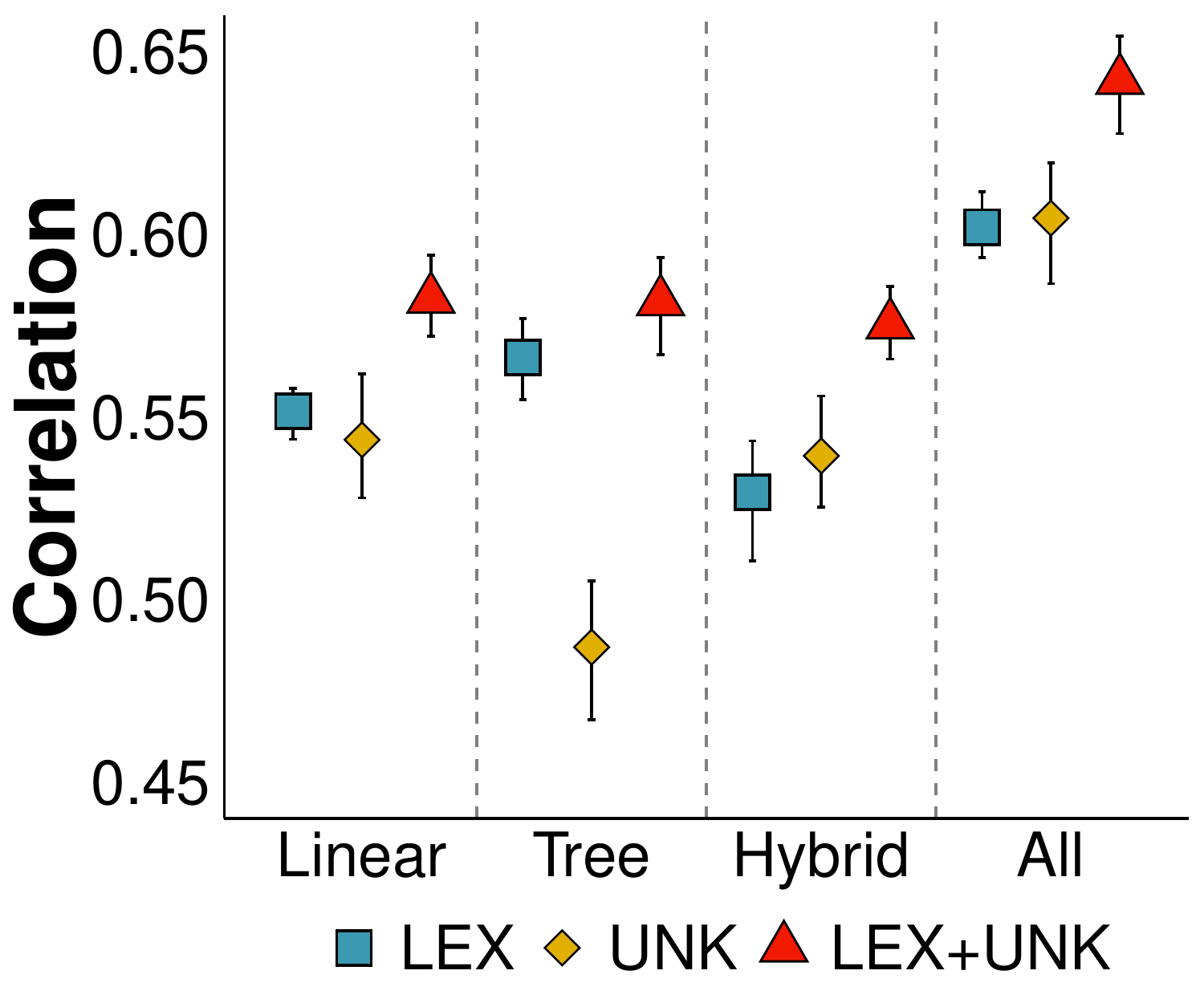}
\vspace{-8mm}
\caption{\small Mean correlation between model predictions and true factuality in nested cross-validation. Error bars show bootstrapped (iter=1,000) 95\% confidence intervals for mean correlation across 10 outer folds.}
\label{fig:correlation}
\vspace{-5mm}
\end{figure}

\figref{fig:correlation} shows the mean correlation between model predictions and true factuality on the outer fold test sets of the nested cross-validation described in \S\ref{sec:model}. We note three aspects of this plot.

First, among the LEX models, the T-biLSTM performs best, followed by the L-biLSTM, then the H-biLSTM. This is somewhat surprising, since \citeauthor{rudinger_neural_2018} find the opposite pattern of performance: the L- and H-biLSTMs vie for dominance, both outperforming the T-biLSTM. This indicates that T-biLSTMs are better able to represent the lexicosyntactic inferences relevant to this dataset, even though they underperform on more general datasets. This possibility is bolstered by the fact that, in contrast to the L- and H-biLSTMs, the LEX version of the T-biLSTMs performs significantly better than the UNK version, suggesting that the T-biLSTM is potentially more reliant on the lexical information than the other two.  

Second, when the LEX and UNK version of each model is ensembled (LEX+UNK), we find comparable performance for all three biLSTMs -- each outperforming the LEX version of the T-biLSTM. This indicates that each model captures similar amounts of information about lexicosyntactic inference, but this information is captured in the models' parameterizations in different ways.

Finally, when all three models are ensembled, we find that both the LEX and UNK version perform significantly better than any specific LEX+UNK model. This may indicate two things: (i) the models that only have access to syntax can perform just as well as ones that have access to both lexical information and syntax; but (ii) these models appear to capture different aspects of inference, since an ensemble of all models (All-LEX+UNK) performs significantly better than either the All-LEX or All-UNK ensembles alone. 

Interestingly, however, even this ensemble performs more than 10 points worse than each model alone on FactBank, UW, and UDS. This raises the question of which lexicosyntactic inferences these models are missing -- investigated below.

\vspace{-1mm}

\section{Analysis}
\label{sec:analysis}

\vspace{-1mm}

We investigate two questions: (i) which inferences do all models do poorly on?; and (ii) what drives the differing strengths of each model?

\begin{table}[t]
\scriptsize
\centering
\begin{tabular}{lrr}
\toprule
                        Someone ... &  True &  Pred. \\
\midrule
           faked that something happened &       -3.15 &       0.86 \\
 was misinformed that something happened &       -2.62 &       1.37 \\
                neglected to do something &       -3.07 &      -0.02 \\
              pretended to have something &       -2.96 &       0.05 \\
          was misjudged to have something &       -2.46 &       0.55 \\
                 forgot to have something &       -3.18 &      -0.17 \\
              neglected to have something &       -2.93 &       0.07 \\
        pretended that something happened &       -2.11 &       0.86 \\
                 declined to do something  &       -3.18 &      -0.22 \\
              was refused to do something  &       -3.16 &      -0.22 \\
                  refused to do something  &       -3.12 &      -0.20 \\
                pretended to do something  &       -3.02 &      -0.11 \\
       disallowed someone to do something  &       -2.56 &       0.34 \\
           was declined to have something  &       -2.36 &       0.55 \\
               declined to have something  &       -3.12 &      -0.23 \\
       did n't hesitate to have something  &        1.84 &      -0.96 \\
                 ceased to have something  &       -2.22 &       0.57 \\
         did n't hesitate to do something  &        1.86 &      -0.92 \\
             lied that something happened  &       -1.99 &       0.78 \\
                feigned to have something  &       -3.07 &      -0.31 \\
\bottomrule
\end{tabular}
\vspace{-2mm}
\caption{\small Sentences with the highest prediction errors.}
\label{tab:badpredictions}
\vspace{-5mm}
\end{table}

\vspace{1mm}

\noindent \textbf{Where do all models fail?} \tabref{tab:badpredictions} shows the 20 sentences with the highest prediction errors under the All-LEX+UNK ensemble. There are two interesting things to note about these sentences. First, most of them involve negative lexicosyntactic inferences that the model predicts to be either positive or near zero. Second, when the true inference is not positive, the matrix polarity of the original sentence is negative. This suggests that the models are not able to capture inferences whose polarity mismatches the matrix clause polarity. 

\begin{figure}[t]
\centering
\includegraphics[scale=.5]{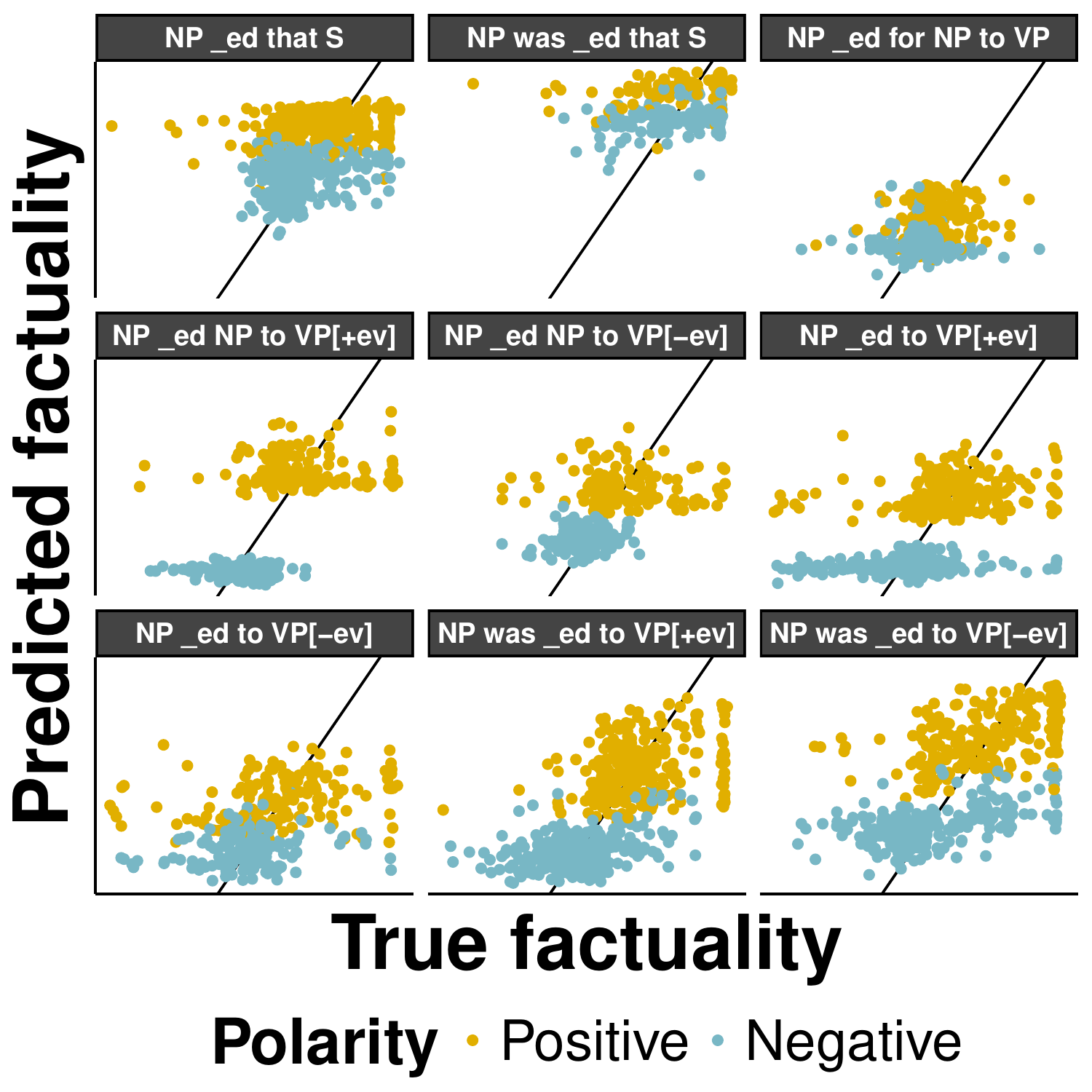}
\vspace{-9mm}
\caption{ Factuality by syntactic context and polarity, each point a verb. Diagonals show perfect prediction.
}
\vspace{-5mm}
\label{fig:truepredicted}
\end{figure}

One question that arises here is whether this inability affects all contexts equally. To answer this, we regress the absolute error of the predictions from this same ensemble (logged and standardized) against true factuality, matrix polarity, and context (as well as all of their two- and three-way interactions).\footnote{See Appendix \ref{sec:regression} for further details, including a summary of the regression on which the above discussion is based.} We find that the three-way interactions in this regression are reliable ($\chi^2$(8)=27.97, $p<$ 0.001) -- suggesting that there are nontrivial differences in these state-of-the-art factuality systems' ability to capture inferential interactions across verbs and syntactic contexts. The differences can be verified visually in \figref{fig:truepredicted}, which plots the factuality predicted by this ensemble against the true factuality from MegaVeridicality2.

To elaborate, the ensemble does best overall on contexts like \ref{ex:forto_analysis} and \ref{ex:acistative_analysis}, and worst overall on contexts like \ref{ex:acistativepassive_analysis}. The contrast between \ref{ex:acistative_analysis} and \ref{ex:acistativepassive_analysis} is particularly interesting because (i) \ref{ex:acistativepassive_analysis} is just the passivized form of \ref{ex:acistative_analysis}; and (ii) we do not observe similar behavior for contexts \ref{ex:acieventive_analysis} and \ref{ex:acieventivepassive_analysis}, which are analogous to \ref{ex:acistative_analysis} and \ref{ex:acistativepassive_analysis}, but replace the stative \textit{have} with the eventive \textit{do}. 

\ex. Someone...
\a. \{\_ed, didn't \_\} for something to happen.\label{ex:forto_analysis}
\b. \{\_ed, didn't \_\} someone to have something.\label{ex:acistative_analysis}
\b. \{was \_ed, wasn't \_ed\} to have something. \label{ex:acistativepassive_analysis} 
\b. \{\_ed, didn't \_\} someone to do something. \label{ex:acieventive_analysis} 
\b. \{was \_ed, wasn't \_ed\} to do something. \label{ex:acieventivepassive_analysis}
\b.  \{\_ed, didn't \_\} that something happened.\label{ex:thatactive_analysis}

An additional nuance is that the ensemble does reliably better on the negative matrix polarity version of \ref{ex:acistative_analysis} than on the positive, with the opposite true for \ref{ex:acieventivepassive_analysis}. This suggests these models do not capture an important inferential interaction between passivization and eventivity.

This suggestion is further bolstered by the fact that the ensemble's ability to predict cases where the matrix polarity mismatches the true factuality are reliably poorer in context \ref{ex:acistativepassive_analysis} but not in its minimal pairs \ref{ex:acieventivepassive_analysis} and \ref{ex:acistative_analysis}, where the ensemble performs reliably poorer when the two match. Indeed, it is contexts \ref{ex:acistativepassive_analysis} and \ref{ex:thatactive_analysis} that drive the polarity mismatch effect evident in \tabref{tab:badpredictions}.

\begin{figure}[t]
\centering
\includegraphics[scale=.5]{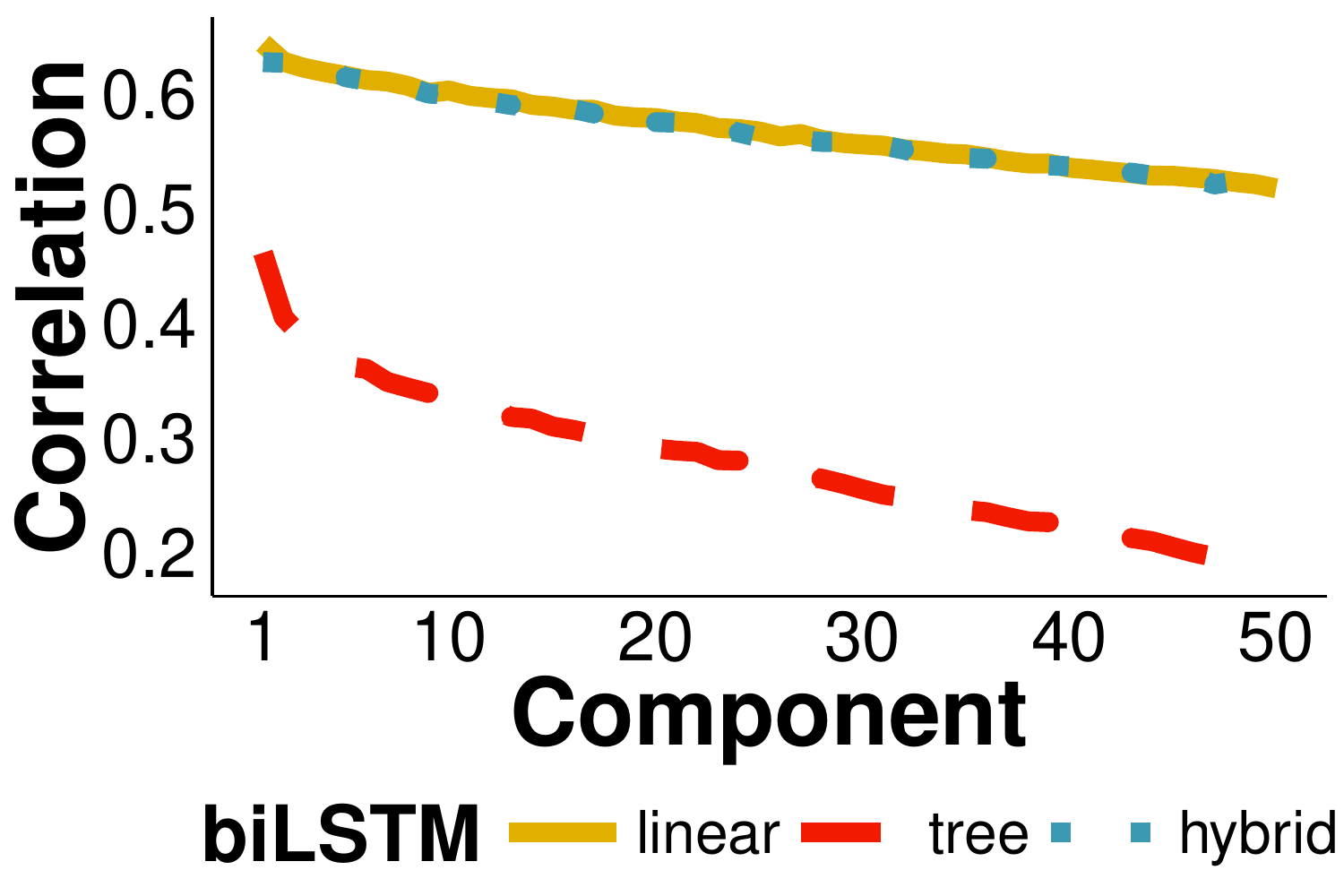}
\vspace{-9mm}
\caption{Canonical correlations between embedding verb embeddings and embedded verb hidden states.}
\vspace{-5mm}
\label{fig:canonicalcorrelation}
\end{figure}

\vspace{1mm}

\noindent \textbf{What drives differences between models?} In \S\ref{sec:results}, we noted two ways that the biLSTMs we investigate differ: (i) the T-biLSTM appears to be more reliant on lexical information than L- and H-biLSTMs; and (ii) each model appears to encode information about lexicosyntactic inference in its parameterizations in different ways. We hypothesize that these two differences are related -- specifically, that the T-biLSTM's heavier reliance on lexical information comes about as a consequence of stronger entanglement between lexical and syntactic information in its hidden states.  

To probe this, we ask to what extent the embedding verb's embedding can be recovered from the embedded verb's hidden state using linear functions. If the lexical information is more strongly entangled with the syntactic information, it should be more difficult to construct a homomorphic (linear) function to decode the embedding verb's embedding from the embedded verb's hidden state.    
To measure this, we conduct a Canonical Correlation Analysis \citep[CCA;][]{hotelling_relations_1936} between these two vector space representations for every sentence in our dataset. Given two matrices $\Xb$ (the embedding verb embeddings column stacked) and $\Yb$ (the embedded verb hidden states column stacked), CCA constructs matrices $\Ab$ and $\Bb$, such that $\ab_i,\bbb_i = \text{arg}_{\ab',\bbb'}\text{max}\;\text{corr}(\ab'\Xb, \bbb'\Yb)$ and $\text{corr}(\ab_i\Xb, \ab_j\Xb) = \text{corr}(\bbb_i\Yb, \bbb_j\Yb) = 0, \forall i < j$. This guarantees that the canonical correlation at component $i$, $\text{corr}(\ab_i\Xb, \bbb_i\Yb)$, is nonincreasing in $i$, and thus the linearly decodable information about $\Yb$ in $\Xb$ can be assessed using this function. 

\figref{fig:canonicalcorrelation} plots the canonical correlations for the first 50 components for each of the biLSTMs we investigated. We find that the canonical correlations associated with the T-biLSTM are substantially lower than those associated with the L- and H-biLSTMs across these first 50 components. This suggests that the T-biLSTM more strongly entangles lexical and syntactic information, perhaps explaining its apparently heavier reliance on lexical information, observed in \S\ref{sec:results}.

Of note here is that the pattern seen in \figref{fig:canonicalcorrelation} is probably at least partly a consequence of the different nonlinearities used for the L-biLSTM (tanh) and T-biLSTM (ReLU), and not the architectures themselves. But whether or not this pattern is due to the architectures, nonlinearities, or both, the entanglement hypothesis may still help explain the pattern of results discussed in \S\ref{sec:results}.

\vspace{-1mm}

\section{Related work}
\label{sec:relatedwork}

\vspace{-1mm}

This work is inspired by recent work in \textit{recasting} various semantic annotations into natural language inference (NLI) datasets \citep{white_inference_2017,poliak_evaluation_2018,poliak_collecting_2018,wang_glue:_2018} to gain a better understanding of which phenomena standard neural NLI models \citep{bowman_large_2015,conneau_supervised_2017} can capture -- a line of work with deep roots \citep{cooper_using_1996}. The experimental setup -- specifically, the idea of UNKing the embedding verb -- was inspired by recent work that uses hypothesis-only baselines for a similar purpose \citep{gururangan_annotation_2018,poliak_hypothesis_2018,tsuchiya_performance_2018}. This work is also related to the broader investigation of sentence representations -- particularly, tasks aimed at probing these representations' content \citep{pavlick_most_2016,adi_fine-grained_2016,conneau_what_2018,conneau_senteval:_2018,dasgupta_evaluating_2018}.

\section{Conclusion}
\label{sec:conclusion}

\vspace{-1mm}

We investigated neural models' ability to capture lexicosyntactic inference, taking the task of event factuality prediction (EFP) as a case study. We built a factuality judgment dataset for all English clause-embedding verbs in various syntactic contexts and used this dataset to probe current state-of-the-art EFP systems. We showed that these systems make certain systematic errors that are clearly visible through the lens of factuality.

\vspace{-1mm}

\section*{Acknowledgments}
This research was supported by the JHU HLTCOE, DARPA LORELEI and AIDA, NSF-BCS (1748969/1749025), and NSF-GRFP (1232825).  The U.S. Government is authorized to reproduce and distribute reprints for Governmental purposes. The views and conclusions contained in this publication are those of the authors and should not be interpreted as representing official policies or endorsements of DARPA or the U.S. Government.

\vspace{-1mm}

\bibliographystyle{acl_natbib_nourl}
\bibliography{Zotero}


\appendix

\section{Data collection}
\label{sec:datacollectionsupplement}

We manipulate two aspects of the subordinate clause in our extension of the MegaVeridicality1 dataset: (i) whether and how an NP embedded subject is introduced; and (ii) whether the embedded clause contains an eventive predicate (\textit{do}, \textit{happen}) or a stative predicate (\textit{have}). 

The first manipulation is known to give rise to different inferential interactions for predicates that take different kinds of infinitival subordinate clauses -- e.g. \textit{remember}, \textit{forget}. For example, while \ref{ex:rememberthatapp}, \ref{ex:notrememberthatapp}, and \ref{ex:remembertoapp} trigger the inference \ref{ex:inferenceknowapp}, \ref{ex:notremembertoapp} triggers the inference \ref{ex:inferenceremembertoapp}. And just a slight tweak to \ref{ex:remembertoapp} and \ref{ex:notremembertoapp} can make these inferences go away completely: neither \ref{ex:rememberaciapp} nor \ref{ex:notrememberaciapp} trigger an inference to either \ref{ex:inferenceknowapp} or \ref{ex:inferenceremembertoapp}.

\ex. \label{ex:rememberfiniteapp}
\a. Jo remembered that Bo left. \label{ex:rememberthatapp}
\b. Jo didn't remember that Bo left. \label{ex:notrememberthatapp}

\ex. \label{ex:remembercontrolapp}
\a. Bo remembered to leave. \label{ex:remembertoapp}
\b. Bo didn't remember to leave. \label{ex:notremembertoapp}

\ex. \label{ex:remembernptoapp}
\a. Jo remembered Bo to have left. \label{ex:rememberaciapp}
\b. Jo didn't remember Bo to have left. \label{ex:notrememberaciapp}

\ex. 
\a. Bo left. \label{ex:inferenceknowapp}
\b. Bo didn't leave. \label{ex:inferenceremembertoapp}

The second manipulation is known to give rise to importantly different temporal interpretations, which also seem to affect factuality judgments \citep{white_factive-implicatives_2014}. For instance, \textit{believe} is generally rated more acceptable in sentences with stative embedded predicates, like \Next[a], and less acceptable in sentences with eventive embedded predicates, like \Next[b]. 

\ex.
\a. Jo believe Mo to be intelligent.
\b. ?Jo believed Mo to run around the park.

This appears to correlate with certain aspects of the temporal interpretation of such sentences \citep{stowell_tense_1982,pesetsky_zero_1991,boskovic_selection_1996,boskovic_syntax_1997,martin_minimalist_1996,martin_null_2001,grano_control_2012,wurmbrand_tense_2014}.

\section{Model and evaluation}
\label{sec:modelsupplement}

We use three models for event factuality prediction proposed by \citet{rudinger_neural_2018}: a stacked bidirectional linear-chain LSTM (L-biLSTM), a stacked bidirectional dependency tree LSTM (T-biLSTM), and a simple ensemble of the two that \citeauthor{rudinger_neural_2018} refer to as a H(ybrid)-biLSTM. We use the two-layer version of these biLSTMs here.

\subsection{Stacked bidirectional linear LSTM}
\label{ssec:lbilstm}

The L-biLSTM we use is a standard extension of the unidirectional linear-chain LSTM \citep{hochreiter_long_1997} by adding the notion of a layer $l\in\{1,\ldots,L\}$ and a direction $d\in\{\rightarrow, \leftarrow\}$ \citep{graves_hybrid_2013,sutskever_sequence_2014,zaremba_learning_2014}.

\vspace{4mm}

{\centering
  $ \displaystyle
    \begin{aligned}
\fb^{(l,d)}_t &= \sigma\left(\Wb^{(l,d)}_\mathrm{f}\left[\hb^{(l,d)}_{\mathbf{prev}_d(t)}; \xb^{(l,d)}_t\right] + \bbb^{(l,d)}_\mathrm{f}\right) \\    
\ib^{(l,d)}_t &= \sigma\left(\Wb^{(l,d)}_\mathrm{i}\left[\hb^{(l,d)}_{\mathbf{prev}_d(t)}; \xb^{(l,d)}_t\right] + \bbb^{(l,d)}_\mathrm{i}\right) \\    
\ob^{(l,d)}_t &= \sigma\left(\Wb^{(l,d)}_\mathrm{o}\left[\hb^{(l,d)}_{\mathbf{prev}_d(t)}; \xb^{(l,d)}_t\right] + \bbb^{(l,d)}_\mathrm{o}\right) \\    
\hat{\cb}^{(l,d)}_t &= g\left(\Wb^{(l,d)}_\mathrm{c}\left[\hb^{(l,d)}_{\mathbf{prev}_d(t)}; \xb^{(l,d)}_t\right] + \bbb^{(l,d)}_\mathrm{c}\right) \\
\cb^{(l,d)}_t &= \ib^{(l,d)}_t \circ \hat{\cb}^{(l,d)}_t + \fb^{(l,d)}_t \circ \cb^{(l,d)}_{\mathbf{prev}_d(t)} \\
\hb^{(l,d)}_t &= \ob^{(l,d)}_t \circ g\left(\cb^{(l,d)}_t\right) \\
\end{aligned}
$
\par}

\vspace{4mm}

\noindent where $\circ$ is the Hadamard product; $\mathbf{prev}_\rightarrow(t) = t-1$ and $\mathbf{prev}_\leftarrow(t) = t+1$, and $\xb^{(l,d)}_t = \xb_t$ if $l=1$; and $\xb^{(l,d)}_t = [\hb^{(l-1,\rightarrow)}_t;\hb^{(l-1,\leftarrow)}_t]$ otherwise. We follow \citeauthor{rudinger_neural_2018} in setting $g$ to the pointwise nonlinearity tanh.

\subsection{Stacked bidirectional tree LSTM}
\label{ssec:tbilstm}

\citet{rudinger_neural_2018} propose a stacked bidirectional extension to the child-sum dependency tree LSTM \citep[T-LSTM;][]{tai_improved_2015}. The T-LSTM redefines $\mathbf{prev}_\rightarrow(t)$ to return the set of indices that correspond to the children of $w_t$ in some dependency tree. In the case of multiple children one defines $\fb_{tk}$ for each child index $k \in \mathbf{prev}_\rightarrow(t)$ in a way analogous to the equations in \S\ref{ssec:lbilstm} -- i.e. as though each child were the only child -- and then sums across $k$ within the equations for $\ib_t$, $\ob_t$, $\hat{\cb}_t$, $\cb_t$, and $\hb_t$.

\citeauthor{rudinger_neural_2018}'s stacked bidirectional T-biLSTM extends the T-LSTM with a \textit{downward} computation in terms of a $\mathbf{prev}_\leftarrow(t)$ that returns the set of indices that correspond to the \textit{parents} of $w_t$ in some dependency tree.\footnote{\citet{miwa_end--end_2016} propose a similar extension for constituency trees.} The same method for combining children in the upward computation is then used for combining parents in the downward computation.

\vspace{2mm}

{\centering
  $ \displaystyle
    \begin{aligned} 
\fb^{(l,d)}_{tk} &= \sigma\left(\Wb^{(l,d)}_\mathrm{f}\left[\hb^{(l,d)}_k; \xb^{(l,d)}_t\right] + \bbb^{(l,d)}_\mathrm{f}\right) \\
\hat{\hb}^{(l,d)}_t &= \sum_{k\in\mathbf{prev}_d(t)} \hb^{(l,d)}_k\\
\end{aligned}
$
\par}

{\centering
  $ \displaystyle
    \begin{aligned} 
\fb^{(l,d)}_{tk} &= \sigma\left(\Wb^{(l,d)}_\mathrm{f}\left[\hb^{(l,d)}_k; \xb^{(l,d)}_t\right] + \bbb^{(l,d)}_\mathrm{f}\right) \\
\hat{\hb}^{(l,d)}_t &= \sum_{k\in\mathbf{prev}_d(t)} \hb^{(l,d)}_k\\
\ib^{(l,d)}_t &= \sigma\left(\Wb^{(l,d)}_\mathrm{i}\left[\hat{\hb}^{(l,d)}_t; \xb^{(l,d)}_t\right] + \bbb^{(l,d)}_\mathrm{i}\right) \\
\ob^{(l,d)}_t &= \sigma\left(\Wb^{(l,d)}_\mathrm{o}\left[\hat{\hb}^{(l,d)}_t; \xb^{(l,d)}_t\right] + \bbb^{(l,d)}_\mathrm{o}\right) \\
\hat{\cb}^{(l,d)}_t &= g\left(\Wb^{(l,d)}_\mathrm{c}\left[\hat{\hb}^{(l,d)}_t; \xb^{(l,d)}_t\right] + \bbb^{(l,d)}_\mathrm{c}\right) \\
\cb^{(l,d)}_t &= \ib^{(l,d)}_t \circ \hat{\cb}^{(l,d)}_t + \sum_{k\in\mathbf{prev}_d(t)} \fb^{(l,d)}_{tk} \circ \cb^{(l,d)}_k \\
\hb^{(l,d)}_t &= \ob^{(l,d)}_t \circ g\left(\cb^{(l,d)}_t\right) \\
\end{aligned}
$
\par}
\vspace{2mm}

\noindent We follow \citeauthor{rudinger_neural_2018} in using a ReLU pointwise nonlinearity for $g$, and in contrast to other dependency tree-structured T-LSTMs \citep{socher_grounded_2014,iyyer_neural_2014}, not using the dependency labels in any way to make the L- and T-biLSTMs as comparable as possible.

\subsection{Regression model}
\label{ssec:regression}

To predict the factuality $v_t$ for the event referred to by a word $w_t$, we follow \citet{rudinger_neural_2018} in using the hidden states from the final layer of the stacked L- or T-biLSTM as the input to a two-layer regression model.

\vspace{1.5mm}
{\centering
  $ \displaystyle
    \begin{aligned} 
\hb^{(L)}_t &= [\hb^{(L,\rightarrow)}_t; \hb^{(L,\leftarrow)}_t]\\
\hat{v_t} &= \Vb_2\;g\left(\Vb_1\hb^{(L)}_t + \bbb_1\right) + \bbb_2
\end{aligned}
$
\par}
\vspace{1mm}

\noindent where $\hat{v_t}$ is passed to a loss function $\mathbb{L}(\hat{v_t}, v_t)$. we follow \citet{rudinger_neural_2018} in using smooth L1 for $\mathbb{L}$ and a ReLU pointwise nonlinearity for $g$.

We also use the simple ensemble method proposed by \citet{rudinger_neural_2018}, which they call the H(ybrid)-biLSTM. In this hybrid, the hidden states from the final layers of both the stacked L-biLSTM and the stacked T-biLSTM are concatenated and passed through the same two-layer regression model \citep[cf.][]{miwa_end--end_2016,bowman_fast_2016}.

\subsection{Out of vocabulary}
\label{ssec:oov}

We use the same UNKing method used by \citet{rudinger_neural_2018}: a single UNK vector is randomly generated at train time, and all OOV items are mapped to it. For the UNK models described in \S\ref{sec:model}, we map all the embedding verbs to this vector at test.

\subsection{Ensemble model}
\label{ssec:ensemble}

We use a ridge regression to ensemble the predictions from various models. The regularization hyperparameter was tuned in the inner fold of the nested cross-validation described in \S\ref{sec:model} using exhaustive grid search over $\lambda\in\{0.0001, 0.001, 0.01, 0.1, 1., 2., 5., 10., 100.\}$.

\section{Regression analysis}
\label{sec:regression}

We regress the absolute error of the predictions from the All-LEX+UNK ensemble (logged and standardized) against true factuality, matrix polarity, and frame (as well as all of their two- and three-way interactions) using a linear mixed effects model with random intercepts for verb and by-verb random slopes for matrix polarity. \tabref{tab:regression} summarizes the fixed effect coefficients based on a sum coding of matrix polarity (negative = -1, positive = 1) and context (NP was \_ed that S = -1).

\begin{table}[h]
{\scriptsize
\begin{tabular}{lrrr}
\toprule
\multicolumn{1}{l}{}&\multicolumn{1}{c}{Coef $\beta$}&\multicolumn{1}{c}{SE($\beta$)}&\multicolumn{1}{c}{\textbf{t}}\tabularnewline
\midrule
(Intercept)&$ 0.00$&$0.03$&$ 0.1$\tabularnewline
polarity&$ 0.15$&$0.02$&$ 6.2$\tabularnewline
factuality&$ 0.00$&$0.03$&$ 0.1$\tabularnewline
NP \_ed to VP[+ev]&$-0.07$&$0.05$&$-1.3$\tabularnewline
NP \_ed to VP[-ev]&$-0.04$&$0.06$&$-0.6$\tabularnewline
NP was \_ed to VP[+ev]&$ 0.02$&$0.05$&$ 0.3$\tabularnewline
NP was \_ed to VP[-ev]&$ 0.23$&$0.05$&$ 4.7$\tabularnewline
NP \_ed NP to VP[+ev]&$-0.01$&$0.07$&$-0.1$\tabularnewline
NP \_ed NP to VP[-ev]&$-0.30$&$0.08$&$-3.8$\tabularnewline
NP \_ed for NP to VP&$-0.34$&$0.07$&$-5.2$\tabularnewline
NP \_ed that S&$ 0.09$&$0.04$&$ 2.1$\tabularnewline
polarity:factuality&$ 0.02$&$0.03$&$ 0.7$\tabularnewline
polarity:NP \_ed to VP[+ev]&$-0.05$&$0.05$&$-0.8$\tabularnewline
polarity:NP \_ed to VP[-ev]&$ 0.03$&$0.06$&$ 0.4$\tabularnewline
polarity:NP was \_ed to VP[+ev]&$-0.20$&$0.05$&$-4.0$\tabularnewline
polarity:NP was \_ed to VP[-ev]&$-0.09$&$0.05$&$-1.8$\tabularnewline
polarity:NP \_ed NP to VP[+ev]&$-0.06$&$0.07$&$-0.8$\tabularnewline
polarity:NP \_ed NP to VP[-ev]&$ 0.28$&$0.08$&$ 3.4$\tabularnewline
polarity:NP \_ed for NP to VP&$ 0.01$&$0.07$&$ 0.1$\tabularnewline
polarity:NP \_ed that S&$ 0.08$&$0.04$&$ 1.8$\tabularnewline
factuality:NP \_ed to VP[+ev]&$-0.04$&$0.05$&$-0.9$\tabularnewline
factuality:NP \_ed to VP[-ev]&$-0.04$&$0.06$&$-0.7$\tabularnewline
factuality:NP was \_ed to VP[+ev]&$ 0.09$&$0.05$&$ 1.7$\tabularnewline
factuality:NP was \_ed to VP[-ev]&$ 0.06$&$0.05$&$ 1.2$\tabularnewline
factuality:NP \_ed NP to VP[+ev]&$ 0.17$&$0.08$&$ 2.1$\tabularnewline
factuality:NP \_ed NP to VP[-ev]&$-0.18$&$0.10$&$-1.7$\tabularnewline
factuality:NP \_ed for NP to VP&$ 0.13$&$0.10$&$ 1.3$\tabularnewline
factuality:NP \_ed that S&$ 0.03$&$0.05$&$ 0.5$\tabularnewline
polarity:factuality:NP \_ed to VP[+ev]&$ 0.06$&$0.05$&$ 1.3$\tabularnewline
polarity:factuality:NP \_ed to VP[-ev]&$ 0.02$&$0.06$&$ 0.3$\tabularnewline
polarity:factuality:NP was \_ed to VP[+ev]&$ 0.07$&$0.05$&$ 1.4$\tabularnewline
polarity:factuality:NP was \_ed to VP[-ev]&$-0.14$&$0.05$&$-3.0$\tabularnewline
polarity:factuality:NP \_ed NP to VP[+ev]&$-0.05$&$0.08$&$-0.6$\tabularnewline
polarity:factuality:NP \_ed NP to VP[-ev]&$ 0.28$&$0.10$&$ 2.7$\tabularnewline
polarity:factuality:NP \_ed for NP to VP&$ 0.12$&$0.10$&$ 1.2$\tabularnewline
polarity:factuality:NP \_ed that S&$-0.17$&$0.05$&$-3.2$\tabularnewline
\bottomrule
\end{tabular}
}
\vspace{-5mm}
\caption{Fixed effects from regression analysis}
\label{tab:regression}
\vspace{-2mm}
\end{table}

The estimated standard deviation for the verb random intercepts is 0.30, and the estimated standard deviation for the by-verb random slopes for polarity is 0.22. Their estimated correlation between the two is 0.30. The marginal $R^2$ is 0.05 and the conditional $R^2$ is 0.20 \citep{nakagawa_general_2013}.

\end{document}